%% file: acl_latex.tex
\pdfoutput=1

\documentclass[11pt]{article}

\usepackage[final]{acl}

\usepackage{times}
\usepackage{latexsym}
\usepackage{amsmath,amssymb,amsthm,bm}
\usepackage{enumitem}
\usepackage{geometry}
\usepackage{subcaption}
\usepackage{float}
\definecolor{DGray}{gray}{0.7}

\usepackage[T1]{fontenc}

\usepackage[utf8]{inputenc}

\usepackage{microtype}

\usepackage{inconsolata}

\usepackage{hyperref}

\usepackage{graphicx}

\usepackage{wrapfig}
\newcommand{\cD}{\mathcal{D}}
\newcommand{\cC}{\mathcal{C}}

\usepackage[sort,space]{cite}
\usepackage{amsmath,amssymb,amsfonts}
\usepackage{algorithmic}
\usepackage{textcomp}
\usepackage{xcolor}
\usepackage[ruled,linesnumbered]{algorithm2e}
\usepackage{multirow}
\usepackage{makecell}
\usepackage{caption}
\usepackage{booktabs}
\usepackage{outlines}
\usepackage{bbold}
\usepackage{graphicx}
\usepackage{enumitem}
\usepackage{soul}
\usepackage{color}
\usepackage{comment}
\usepackage{diagbox}
\usepackage{epstopdf}
\usepackage{wrapfig}
\usepackage{pifont}
\usepackage{acronym}
\usepackage[most]{tcolorbox}   
\tcbuselibrary{breakable}        
\usepackage{subcaption}
\title{\textsc{Cycle-Instruct}: Fully Seed-Free Instruction Tuning via Dual Self-Training and Cycle Consistency}

\author{Zhanming Shen\textsuperscript{$\spadesuit$}, \ Hao Chen\textsuperscript{$\spadesuit$}, \ Yulei Tang\textsuperscript{$\clubsuit$}, \ Shaolin Zhu\textsuperscript{$\heartsuit$}\\
\ \textbf{Wentao Ye\textsuperscript{$\spadesuit$}}, \ \textbf{Xiaomeng Hu\textsuperscript{$\spadesuit$}}, \ \textbf{Haobo Wang\textsuperscript{$\spadesuit$}}, \ \textbf{Gang Chen\textsuperscript{$\spadesuit$}}, \ \textbf{Junbo Zhao\textsuperscript{$\spadesuit$}\thanks{Corresponding author}} \\
  \textsuperscript{$\spadesuit$}Zhejiang University \quad
  \textsuperscript{$\clubsuit$}University of Science and Technology of China \quad
  \textsuperscript{$\heartsuit$}Tianjin University \\
  \texttt{\{z.shen, j.zhao\}@zju.edu.cn}
}

\begin{document}
\maketitle
\begin{abstract}
Instruction tuning is vital for aligning large language models (LLMs) with human intent, but current methods typically rely on costly human-annotated seed data or powerful external teacher models. 
While instruction back-translation techniques reduce this dependency, they remain fundamentally tethered to an initial seed set, which limits full automation, introduces biases, and can lead to inefficient use of unlabeled corpora. 
In this paper, we propose \textsc{Cycle-Instruct}, a novel framework that achieves fully seed-free instruction tuning. 
Inspired by cycle consistency, \textsc{Cycle-Instruct} employs a dual self-training loop where two models—an answer generator and a question generator—are bootstrapped solely from raw, unlabeled text. 
These models mutually supervise each other by reconstructing original text segments from their counterpart's generated pseudo-labels, effectively learning from the intrinsic structure of the data without any human-provided seeds. 
We demonstrate \textsc{Cycle-Instruct}'s efficacy across four diverse data tracks, including general instruction-following, domain-specific tasks, dialogue logs, and plain text. 
Our extensive experiments show that \textsc{Cycle-Instruct} not only outperforms seed-driven back-translation baselines but also achieves performance comparable to strongly supervised methods. 
\end{abstract}

\input{latex/sections/intro}
\input{latex/sections/siplfied_background}

\input{latex/sections/method}

\input{latex/sections/experiment}

\section{Conclusion}

We presented \textsc{Cycle-Instruct}, a fully seed-free instruction-tuning framework that relies on neither human-written seeds nor external teacher models.  Through a dual self-training loop—an answer generator and a question generator linked by cycle consistency—the method extracts high-quality instruction–response pairs straight from raw text.  Extensive experiments show that \textsc{Cycle-Instruct} consistently outperforms seed-driven back-translation baselines and delivers performance competitive with models trained on strong supervised data, underscoring its ability to tap vast unlabeled corpora for scalable, automated alignment of LLMs with human intent.

\clearpage

\section*{Limitations}

First, due to resource constraints we only fine-tune model with LoRA, so the approach’s behaviour at full-parameter or larger scales remains untested; second, the cycle-consistency objective could be exploited to reconstruct user prompts, posing potential privacy risks that call for defences such as differential privacy or rigorous red-teaming before deployment; finally, our seed-free segmentation relies on the presence of a question mark, which can fail on narrative or expository texts without explicit interrogatives, suggesting that richer discourse cues or retrieval-based heuristics should be explored.

\bibliography{custom}

\appendix
\input{latex/sections/appendix}

\end{document}

%% file: latex/sections/intro.tex
\section{Introduction}

Instruction tuning \citep{ouyang2022training} has emerged as a crucial technique for aligning large language models (LLMs) with human intent, enabling effective generalization across diverse instruction-based tasks \citep{ouyang2022training,wei2021finetuned,touvron2023llama}.
However, conventional instruction tuning typically requires extensive human-annotated data \citep{kopf2304openassistant,conover2023free} or relies on powerful external teacher models \citep{taori2023stanford,yin2023dynosaur}. 
These dependencies are not only costly and limit scalability \citep{wang2022self} but are also often inapplicable in certain settings—for example, privacy-preserving scenarios \citep{zhang2024towards}.

Recent advances in \emph{instruction back-translation} \citep{li2023self,koksal2023longform} have sought to mitigate above issues by leveraging unlabeled text. 
These methods typically train a model on a small seed set of (instruction, answer) pairs to generate candidate instructions for unlabeled documents, which are then filtered to create additional training examples \citep{chen2024reinstruct,chen2023dog}. 
While reducing reliance on extensive human labeling, these back-translation pipelines still critically hinge on an initial, manually-curated seed set.

\begin{figure}[t]
  \centering
  \includegraphics[width=1.0\linewidth]{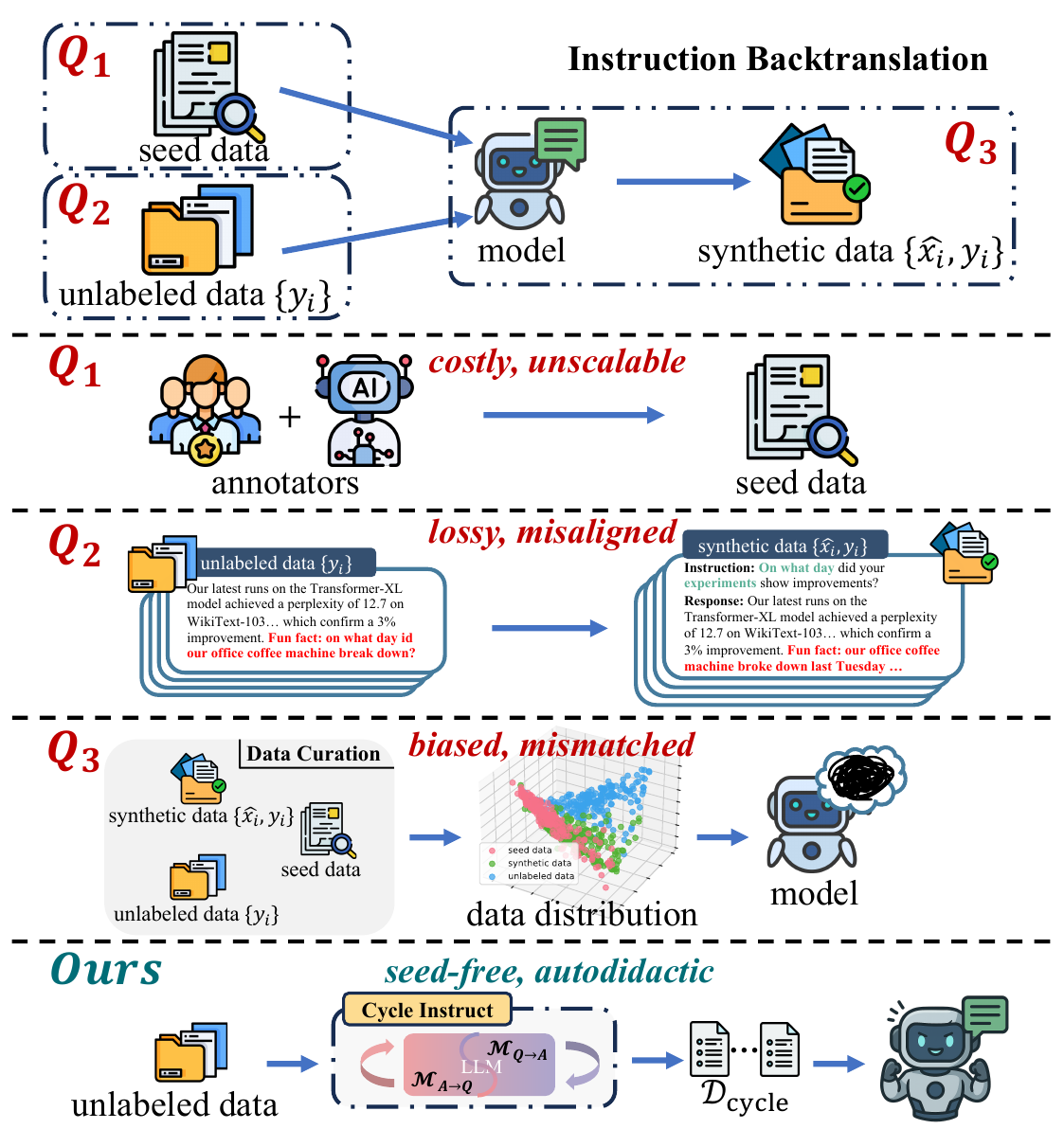}
  \caption{\textbf{Seed-dependency bottleneck:}  
(1) \emph{Costly seed curation;}  
(2) \emph{Data wastage \& mis-paired;}
(3) \emph{Bias transfer \& low diversity.}}
  \label{fig:seed-dependency}
\end{figure}

This seed-dependency causes a core challenge: \textbf{the reliance on seed data inherently limits the full automation, diversity, and data efficiency of instruction tuning}. 
Specifically, as illustrated in Figure~\ref{fig:seed-dependency}, (i) assembling a diverse, high-quality seed set remains labour-intensive and costly; 
(ii) conditioning generation on a small seed corpus can transfer its stylistic and topical biases to the synthetic pairs, curbing diversity and generalisation; and (iii) the common practice of treating all unlabeled passages as answers can lead to data wastage, as question-formatted segments in raw corpora are often discarded or receive low-quality synthetic instructions. 
This raises a critical question: \emph{How can we unlock the full potential of abundant raw text for instruction tuning, without the bottleneck of seed data, while ensuring high-quality and diverse instruction-following capabilities?}

To address this challenge, we propose \textsc{Cycle-Instruct}, a novel and fully \textbf{seed-free} framework for instruction tuning that requires no manually written seeds or external teacher models. 
Inspired by cycle consistency in unsupervised machine translation \citep{he2016dual,lample2018phrase}, 
\textsc{Cycle-Instruct} employs a dual self-training loop. 
We begin by automatically partitioning a large unlabeled corpus into \emph{potential question passages} and \emph{potential answer passages}, reformatted into instruction-tuning compliant \texttt{<instruction>} and \texttt{<response>} slots. 
An answer generator produces pseudo-responses for question passages, while a question generator back-translates pseudo-instructions for answer passages. 
These two models then act as mutual teachers: each learns to reconstruct the original passage by conditioning on the pseudo-label from its counterpart, using the reconstruction error as the training objective. 
Crucially, because every sample is supervised by its own \textbf{ground-truth content}, the learning signal closely approximates fully supervised training, allowing the models to internalise the true data distribution from the entire unlabeled corpus.

We thoroughly evaluate \textsc{Cycle-Instruct} across four diverse tracks.
Our results consistently demonstrate that \textsc{Cycle-Instruct} generates coherent and relevant instruction-following data, establishing its potential as a scalable, especially when human-labeled resources are scarce or unavailable.
Our main contributions are as follows:
\begin{itemize}[leftmargin=*]
    \item We propose \textsc{Cycle-Instruct}, a \emph{fully seed-free} instruction-tuning framework that entirely eliminates reliance on human-written seeds and external teacher models, directly addressing the core bottleneck of current back-translation methods.
    
    \item We introduce a \textbf{dual self-training loop with cycle consistency} as the core mechanism, enabling two models to mutually supervise each other by reconstructing raw passages, yielding a high-quality learning signal akin to fully supervised training from unlabeled text alone.
    
    \item Through extensive experiments on four diverse data tracks, we demonstrate that \textsc{Cycle-Instruct} achieves performance \textbf{comparable to} strong supervised methods and significantly \textbf{outperforms} seed-driven back-translation baselines, all while requiring \emph{zero} human annotations.
\end{itemize}

\begin{figure*}[t]
  \centering
  \includegraphics[width=1.0\textwidth]{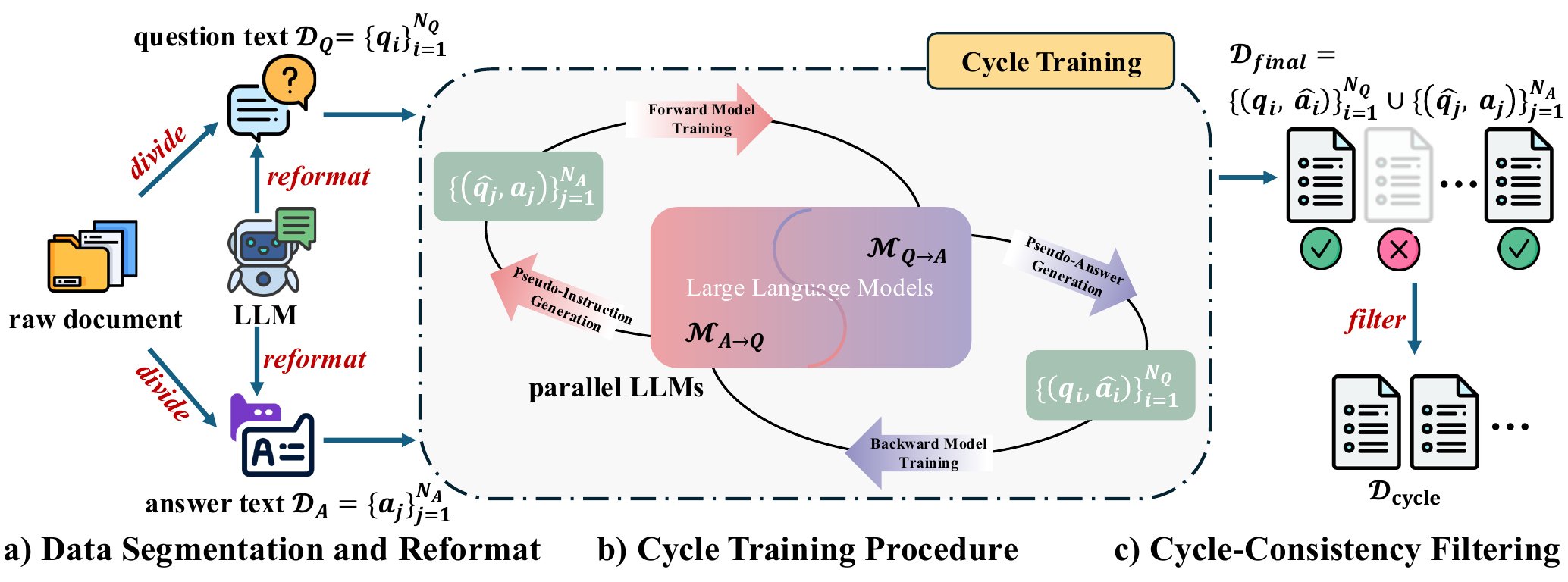}
  \caption{The Overall Framework of our Fully Seed-Free Instruction Tuning Method.}
  \label{fig:workflow}
\end{figure*}

%% file: latex/sections/siplfied_background.tex
\section{Background}
\subsection{Back-Translation for Instruction Tuning}
\vspace{2pt}

Back-translation (BT) was introduced in NMT to exploit monolingual target data by “translating it back’’ into the source language \citep{sennrich2016edinburgh}.  
Recent work \citep{li2023self,koksal2023longform} adapts this idea to instruction tuning.

Let $\mathcal S=\{(q,a)\}$ be a \emph{small seed} set of question–answer pairs and  
$\mathcal D_A$ an unlabeled corpus of free-form answers.  
\vspace{-2pt}
\begin{enumerate}[leftmargin=1.3em]
\item \textbf{Seed training.}  
      Train an \emph{inverse model} $F_{A\!\to\!Q}$ to predict a question from an answer:
      \begin{equation}
        \mathcal L_{\text{inv}}
          \;=\;
          -\!\!\sum_{(q,a)\in\mathcal S}\!\log p_\psi\bigl(q\,\big|\,a\bigr).
      \end{equation}
\item \textbf{Pseudo-pair generation.}  
      For each $a\in\mathcal D_A$, create a pseudo-question
      $\hat q = F_{A\!\to\!Q}(a)$ and form a synthetic pair $(\hat q,a)$.
\item \textbf{BT fine-tuning.}  
      Fine-tune the forward model $G_{Q\!\to\!A}$ on all synthetic pairs via
      \begin{equation}
        \mathcal L_{\text{BT}}
          =\mathbb E_{(\hat q,a)}\!\bigl[-\log p_\theta(a\mid\hat q)\bigr].
      \end{equation}
\end{enumerate}

The resulting model $G_{Q\!\to\!A}$ benefits from a much larger, automatically generated corpus, while all notation remains consistent with the cycle-consistency formulation that follows.

\subsection{Cycle Consistency}
Cycle consistency traces its roots to unsupervised machine translation and dual learning, where two models—one translating from language $X$ to $Y$ and the other from $Y$ to $X$—are trained jointly by enforcing that translating “there and back” recovers the original sentence \citep{he2016dual}.  Concretely, given mappings
\begin{equation}
G: X \to Y,\quad F: Y \to X,
\end{equation}
one adds the reconstruction loss
\begin{equation}
\begin{aligned}
\mathcal{L}_{\mathrm{cycle}}
  &= \mathbb{E}_{x\sim\mathcal{D}_X}
     \!\bigl[\ell\bigl(x,\,F(G(x))\bigr)\bigr] \\
  &\quad
     + \mathbb{E}_{y\sim\mathcal{D}_Y}
       \!\bigl[\ell\bigl(y,\,G(F(y))\bigr)\bigr],
\end{aligned}
\end{equation}

where $\ell(\cdot,\cdot)$ is typically the negative log-likelihood of reconstructing the original sequence. CycleGAN \citep{chu2017cyclegan} applied a similar cycle constraint in image translation by using $\ell_1$ pixel losses, illustrating the concept across modalities.

In Cycle-Instruct, we reinterpret this mechanism for instruction tuning by viewing questions and answers as two analogous “languages.”  Let
\begin{equation}
G_{\!Q\to A}: \mathcal{Q}\to\mathcal{A},\quad
F_{\!A\to Q}: \mathcal{A}\to\mathcal{Q}
\end{equation}
be the question-to-answer and answer-to-question models, respectively.  We then minimize
\begin{equation}
\begin{aligned}
\mathcal{L}_{\mathrm{cycle}}
  &= \mathbb{E}_{q\sim\mathcal{Q}}
     \!\Bigl[\ell\bigl(q,\,
       F_{A\to Q}\!\bigl(G_{Q\to A}(q)\bigr)\bigr)\Bigr] \\
  &\quad
     + \mathbb{E}_{a\sim\mathcal{A}}
       \!\Bigl[\ell\bigl(a,\,
         G_{Q\to A}\!\bigl(F_{A\to Q}(a)\bigr)\bigr)\Bigr].
\end{aligned}
\end{equation}
By enforcing $F_{A\to Q}(G_{Q\to A}(q))\approx q$ and $G_{Q\to A}(F_{A\to Q}(a))\approx a$ on unlabeled text, the two models effectively teach one another.  This self-supervised reconstruction signal enables the generation of high-quality question–answer pairs directly from raw corpora, without any seed examples or external teachers.

%% file: latex/sections/method.tex
\section{Methodology}

\subsection{Overview}
We propose \textsc{Cycle-Instruct}, a seed-free instruction tuning framework that leverages cycle-consistent dual-model training. Unlike previous instruction tuning methods requiring human-written seeds or external teacher models, \textsc{Cycle-Instruct} uses raw unlabeled text to iteratively bootstrap two LLMs: a forward model \(\mathcal{M}_{Q\rightarrow A}\) and a backward model \(\mathcal{M}_{A\rightarrow Q}\). The forward model generates pseudo-responses from extracted question passages, and the backward model back-translates pseudo-instructions from answer passages. Both models mutually reinforce each other's predictions, with training losses computed based on reconstruction of original data segments. Figure ~\ref{fig:workflow} illustrates the overall framework of our
method.

\subsection{Data Segmentation}
We adopt an ultra-light, seed-free rule: a passage is a question iff it
contains at least one question mark “\texttt{?}”; otherwise it is treated
as an answer.

Concretely, each raw document $\cD_{\text{raw}}$ is split into paragraphs
by blank lines, and we obtain the raw paragraph sets
\begin{equation}
  \cD_Q^{\mathrm{raw}}=\bigl\{\,q_i^{\mathrm{raw}}\,\bigr\}_{i=1}^{N_Q},
  \quad
  \cD_A^{\mathrm{raw}}=\bigl\{\,a_j^{\mathrm{raw}}\,\bigr\}_{j=1}^{N_A},
\end{equation}
where $q_i^{\mathrm{raw}}$ (resp.\ $a_j^{\mathrm{raw}}$) is a paragraph
with (resp.\ without) a “\texttt{?}”.

\subsection{Data Reformat}
To turn the split raw passages $\cD_Q^{\mathrm{raw}}$ and $\cD_A^{\mathrm{raw}}$ into \textsc{instruction–response} style data
we apply two fixed rewriting prompts:

\begin{itemize}[leftmargin=*]
  \item \textbf{Prompter (questions).}
        For each $q_i^{\mathrm{raw}}\!\in\!\cD_Q^{\mathrm{raw}}$ we ask
        the model to rewrite the paragraph into one self-contained,
        natural-sounding question $q_i$ (See template in Appendix~\ref{PROMPTER}).
  \item \textbf{Assistant (answers).}
        For each $a_j^{\mathrm{raw}}\!\in\!\cD_A^{\mathrm{raw}}$ we ask
        the model to polish the text into a coherent answer paragraph $a_j$ without introducing new information (See template in Appendix~\ref{ASSISTANT}).
\end{itemize}

The rewritten segments constitute the standardized datasets
\begin{equation}
  \cD_Q=\bigl\{\,q_i\,\bigr\}_{i=1}^{N_Q},
  \quad
  \cD_A=\bigl\{\,a_j\,\bigr\}_{j=1}^{N_A},
\end{equation}
providing paired forms $(q_i,~\_)$ and $(\_,~a_j)$ that feed the
four-step cycle training loop described latter.

\subsection{Cycle Training Procedure}
We instantiate two transformer models from the same base model (e.g., LLaMA3):
\begin{itemize}
    \item Forward model: \(\mathcal{M}_{Q\rightarrow A}(q; \theta_{Q\rightarrow A})\) generates responses given instructions.
    \item Backward model: \(\mathcal{M}_{A\rightarrow Q}(a; \theta_{A\rightarrow Q})\) generates instructions given responses.
\end{itemize}

Here, \(\theta_{Q\rightarrow A}\) and \(\theta_{A\rightarrow Q}\) represent the parameters used by each model during generation, which are trained separately. However, under our self-training assumption, both models share the same base architecture.

Training proceeds iteratively in four cyclical steps:

\textbf{Step 1: Pseudo-Answer Generation} (using \(\mathcal{M}_{Q\rightarrow A}\)) :  
Given \(\mathcal{D}_Q\), we generate pseudo-responses \(\hat{a}_i\) (See template in Appendix~\ref{Pseudo-Answer Generation}.):
\begin{equation}
    \hat{a}_i = \mathcal{M}_{Q\rightarrow A}(q_i; \theta_{Q\rightarrow A}), \quad \forall q_i \in \mathcal{D}_Q
\end{equation}

resulting in pseudo-labeled pairs \(\{(q_i, \hat{a}_i)\}_{i=1}^{N_Q}\).

\textbf{Step 2: Backward Model Training} (updating \(\mathcal{M}_{A\rightarrow Q}\)):  
Using pairs \((q_i, \hat{a}_i)\), we minimize the negative log-likelihood of reconstructing the original instructions:
\begin{equation}
    \mathcal{L}_{A\rightarrow Q} = -\frac{1}{N_Q} \sum_{i=1}^{N_Q} \log P(q_i \mid \hat{a}_i; \theta_{A\rightarrow Q})
\end{equation}

\textbf{Step 3: Pseudo-Instruction Generation} (using \(\mathcal{M}_{A\rightarrow Q}\)):  
Given \(\mathcal{D}_A\), we generate pseudo-instructions \(\hat{q}_j\) (See template in Appendix~\ref{Pseudo-Instruction Generation}.):
\begin{equation}
    \hat{q}_j = \mathcal{M}_{A\rightarrow Q}(a_j; \theta_{A\rightarrow Q}), \quad \forall a_j \in \mathcal{D}_A
\end{equation}
producing pseudo-labeled pairs \(\{(\hat{q}_j, a_j)\}_{j=1}^{N_A}\).

\textbf{Step 4: Forward Model Training} (updating \(\mathcal{M}_{Q\rightarrow A}\)):  
Using pairs \((\hat{q}_j, a_j)\), we minimize the negative log-likelihood of reconstructing the original responses:
\begin{equation}
    \mathcal{L}_{Q\rightarrow A} = -\frac{1}{N_A} \sum_{j=1}^{N_A} \log P(a_j \mid \hat{q}_j; \theta_{Q\rightarrow A})
\end{equation}

\textbf{Iterative Refinement}:
We repeat Steps 1–4 iteratively, with each cycle progressively improving pseudo-label quality and better approximating the true underlying distribution of the unlabeled corpus.

\textbf{Final Dataset Construction}:
After completing a fixed number of cycles \(T\), we construct the final synthetic instruction-following dataset:
\begin{equation}
    \mathcal{D}_{\text{final}} = \{(q_i, \hat{a}_i)\}_{i=1}^{N_Q} \cup \{(\hat{q}_j, a_j)\}_{j=1}^{N_A}
\end{equation}
This merged dataset combines pseudo-responses generated by \(\mathcal{M}_{Q\rightarrow A}\) and pseudo-instructions generated by \(\mathcal{M}_{A\rightarrow Q}\).

\begin{table*}[t]
\centering
\small
\setlength{\tabcolsep}{4pt}
\begin{tabular}{l|c|ccccc|ccccc}
\toprule
 & & \multicolumn{5}{c|}{\textbf{Alpaca–GPT4}} & \multicolumn{5}{c}{\textbf{Dolly-15k}} \\
\cmidrule(r){3-7}\cmidrule(l){8-12}
\textbf{Method} & \textbf{Annot.\,(\%)} & MMLU & BBH & CRASS & DROP & \textbf{Avg} & MMLU & BBH & CRASS & DROP & \textbf{Avg}\\
\midrule
Vanilla      & 0   & 55.85 & 37.16 & 59.48 & 36.01 & 47.13 & 55.85 & 37.16 & 59.48 & 36.01 & 47.13\\
\midrule
\multirow{3}{*}{Random} 
             & 5   & 57.92 & 38.38 & 68.98 & 38.06 & 50.84 & 55.42 & 37.14 & 62.41 & 33.79 & 47.19\\
             & 10  & 57.41 & 37.68 & 66.42 & 36.72 & 49.56 & 55.46 & 36.18 & 62.77 & 33.09 & 46.88\\
             & 20  & 57.63 & 37.23 & 67.88 & 37.39 & 50.03 & 55.91 & 36.51 & 63.50 & 33.06 & 47.25\\
\midrule
\multirow{3}{*}{Cluster}
             & 5   & 57.26 & 38.00 & 67.88 & 36.95 & 50.02 & 55.31 & 34.73 & 59.12 & 33.37 & 45.63\\
             & 10  & 57.57 & 37.68 & 68.61 & 36.75 & 50.15 & 55.48 & 35.83 & 61.68 & 33.57 & 46.64\\
             & 20  & 57.78 & 38.34 & 68.61 & 37.61 & 50.59 & 56.87 & 35.66 & 65.69 & 33.89 & 48.03\\
\midrule
\textbf{Cycle-Inst (Ours)}   & 0   & 59.01 & 39.28 & 74.82 & 39.86 & 53.24 & \textbf{58.96} & 37.44 & 70.07 & 37.79 & 51.07\\
\textbf{Cycle-Filt (Ours)}  & 0   & \textbf{59.39} & 39.46 & 77.37 & 40.46 & 54.17 & 58.26 & 37.51 & 70.44 & 37.80 & 51.00\\
\midrule
\textcolor{DGray}{SFT-80}    & \textcolor{DGray}{80}  &
\textcolor{DGray}{58.57} & \textcolor{DGray}{39.68} & \textcolor{DGray}{74.45} & \textcolor{DGray}{39.47} & \textcolor{DGray}{53.04} &
\textcolor{DGray}{58.65} & \textcolor{DGray}{39.04} & \textcolor{DGray}{71.53} & \textcolor{DGray}{40.91} & \textcolor{DGray}{52.53} \\
\textcolor{DGray}{All-SFT}  & \textcolor{DGray}{100} &
\textcolor{DGray}{58.84} & \textcolor{DGray}{\textbf{40.53}} & \textcolor{DGray}{\textbf{79.20}} & \textcolor{DGray}{\textbf{41.37}} & \textcolor{DGray}{\textbf{54.99}} &
\textcolor{DGray}{58.92} & \textcolor{DGray}{\textbf{39.72}} & \textcolor{DGray}{\textbf{74.09}} & \textcolor{DGray}{\textbf{41.05}} & \textcolor{DGray}{\textbf{53.45}} \\
\bottomrule
\end{tabular}
\caption{Results on Alpaca–GPT4 and Dolly-15k for \textbf{Llama-3.1-8B}. Our methods surpass all back-translation baselines and approach the fully supervised \textbf{All-SFT} scores.}

\label{tab:general-final}
\end{table*}

\subsection{Cycle-Consistency Filtering (Optional)}
\label{sec:cycle-consistency}
Although $\cD_{\text{final}}$ is already seed-free, we can further
\emph{audit and prune} its pseudo labels by checking cycle
consistency under the final checkpoints
$\theta_{Q\rightarrow A}^{(T)}$ and
$\theta_{A\rightarrow Q}^{(T)}$.

\paragraph{(1) One-step reconstruction.}
For every pair in $\cD_{\text{final}}$ we pass the \emph{pseudo} side
through the \emph{opposite} model to obtain a reconstructed label:
\begin{equation}
\tilde q_i = M_{A\rightarrow Q}( \hat a_i;\theta_{A\rightarrow Q}^{(T)},\ \ 
\tilde a_j = M_{Q\rightarrow A}( \hat q_j;\theta_{Q\rightarrow A}^{(T)}).
\end{equation}

\paragraph{(2) Embedding distance.}
We encode the original label and its reconstruction with the same
sentence encoder $\phi(\cdot)$ (we use the LLaMA3 inference encoder
without fine-tuning):
\begin{equation}
d_i=\|\phi(q_i)-\phi(\tilde q_i)\|_2,\ \ 
d_j=\|\phi(a_j)-\phi(\tilde a_j)\|_2 .
\end{equation}

\paragraph{(3) Semantic clustering.}
To avoid domain or format bias in later pruning, we cluster the
\emph{gold sides}—$\{q_i\}$ and $\{a_j\}$—in the embedding space via
$k$-means ($k\!=\!200$ by default, tuned on a held-out slice).  Each
cluster thus represents a local region of the original data
distribution.

\paragraph{(4) k-center greedy pruning.}
Within every cluster $\cC$ we apply the \emph{k-center greedy}
selection rule to rank samples by distance score
($d_i$ or $d_j$).  We drop the top $5\%$ farthest points:
\begin{equation}
  \cC_{\text{keep}}=\cC\setminus
  \operatorname{TopPercent}(\cC,\text{dist},5\%).
\end{equation}
Because k-center greedy iteratively adds points that maximise the
minimum pairwise distance, the retained set still covers the semantic
support of $\cC$ while discarding outliers that even the
well-trained opposite model fails to reconstruct faithfully.

\paragraph{Resulting corpus.}
Concatenating pruned clusters yields
\begin{equation}
  \cD_{\text{cycle}} \subset \cD_{\text{final}},
  \qquad
  |\cD_{\text{cycle}}| \approx 0.95\,|\cD_{\text{final}}|.
\end{equation}
We use $\cD_{\text{cycle}}$ for all downstream supervised fine-tuning
experiments.


%% file: latex/sections/experiment.tex
\section{Experimental Setup}
\label{sec:exp-setup}


\subsection{Dataset Tracks and Baselines}
\label{ssec:tracks}

We first test the proposed method on general instruction datasets. Next, we apply it to domain-specific datasets with limited labeled data. Finally and most importantly, following existing backtranslation works \citep{li2023self,koksal2023longform,chen2023dog}, we conduct experiments on dialogue and plain text documents to showcase the effectiveness of our approach in generating coherent instruction-following data from raw text. For data in documents track, we leverage GPT-4o Mini to generate golden labels for the data, using the prompts provided in the Appendix ~\ref{Pseudo-Answer Generation}\&~\ref{Pseudo-Instruction Generation}.

\begin{table*}[t]
\centering
\small
\setlength{\tabcolsep}{4pt}
\begin{tabular}{l|c|cccccccccc}
\toprule
\textbf{Method} & \textbf{Annot.\,(\%)} & CK & CB & CC & CM & HB & HC & MG & PM & \textbf{Avg}\\
\midrule
Vanilla      & 0   & 0.630 & 0.701 & 0.390 & 0.555 & 0.706 & 0.448 & 0.670 & 0.614 & 0.589\\
\midrule
\multirow{3}{*}{Random}
             & 5   & 0.657 & 0.715 & 0.410 & \textbf{0.595} & 0.719 & 0.453 & 0.720 & \textbf{0.665} & 0.617\\
             & 10  & 0.679 & 0.729 & 0.440 & 0.584 & 0.739 & 0.458 & 0.750 & 0.640 & 0.627\\
             & 20  & 0.679 & 0.708 & 0.440 & 0.584 & 0.748 & 0.478 & \textbf{0.760} & 0.662 & 0.632\\
\midrule
\multirow{3}{*}{Cluster}
             & 5   & 0.679 & 0.729 & 0.440 & 0.584 & 0.732 & 0.458 & 0.730 & 0.658 & 0.626\\
             & 10  & 0.657 & 0.736 & 0.430 & 0.566 & 0.745 & 0.468 & 0.750 & 0.629 & 0.623\\
             & 20  & 0.660 & 0.701 & 0.440 & 0.584 & \textbf{0.752} & 0.488 & 0.730 & 0.654 & 0.626\\
\midrule
\textbf{Cycle-Inst (Ours)}   & 0   & 0.668 & 0.715 & \textbf{0.460} & 0.578 & 0.742 & 0.473 & \textbf{0.760} & 0.643 & 0.630\\
\textbf{Cycle-Filt (Ours)}   & 0   & \textbf{0.687} & \textbf{0.757} & \textbf{0.460} & 0.566 & 0.748 & \textbf{0.483} & 0.730 & 0.658 & \textbf{0.636}\\
\midrule
\textcolor{DGray}{SFT-80}    & \textcolor{DGray}{80}  &
\textcolor{DGray}{0.657} & \textcolor{DGray}{0.743} & \textcolor{DGray}{0.440} & \textcolor{DGray}{0.572} &
\textcolor{DGray}{0.726} & \textcolor{DGray}{0.478} & \textcolor{DGray}{0.730} & \textcolor{DGray}{0.654} &
\textcolor{DGray}{0.625} \\
\textcolor{DGray}{All-SFT}  & \textcolor{DGray}{100} &
\textcolor{DGray}{0.679} & \textcolor{DGray}{0.736} & \textcolor{DGray}{0.450} & \textcolor{DGray}{0.566} &
\textcolor{DGray}{0.735} & \textcolor{DGray}{0.468} & \textcolor{DGray}{0.730} & \textcolor{DGray}{0.647} &
\textcolor{DGray}{0.626} \\
\bottomrule
\end{tabular}
\caption{MedAlpaca results on the nine medical sub-domains of \textsc{MMLU} for \textbf{Llama-3.1-8B}. Our seed-free methods outperform every back-translation baseline and even surpass the fully supervised \textbf{All-SFT}.}
\label{tab:medical-final}
\end{table*}

\vspace{1mm}
\paragraph{Track 1 — General instructions.}
Curated corpora such as \textbf{Alpaca–GPT4} \citep{peng2023instruction} and \textbf{Dolly-15k}\footnote{\url{https://huggingface.co/datasets/databricks/databricks-dolly-15k}} provide human or GPT-4-authored instruction datasets, yet one side of
the pair is frequently missing or unreliable.

\paragraph{Track 2 — Domain-specific instructions.}
In specialist domains (e.g.\ medicine) aligned Q–A pairs are scarce.
Using \textbf{Medical-Alpaca}\footnote{\url{https://huggingface.co/datasets/medalpaca/medical_meadow_medical_flashcards}}. we randomly remove one side of
the pair, retaining 20 k unpaired questions and answers—emulating the common situation where
only FAQ-style queries and unlabeled clinical notes exist.

\paragraph{Track 3 — Dialogue logs.}
Conversational logs preserve turn order but not explicit Q–A alignment.
From \textbf{OASST-1}\footnote{\url{https://huggingface.co/datasets/OpenAssistant/oasst1}} we identify questions via a “\texttt{?}”
heuristic (English + Chinese) and treat residual turns as answer
candidates, yielding 15,126 Q and 24,874 A fragments.

\paragraph{Track 4 — Plain text.}
Narrative corpora embed latent questions within paragraphs.
From 40 k \textbf{WikiHow}\footnote{\url{https://huggingface.co/datasets/wangwilliamyang/wikihow}} articles we extract 5,178 interrogatives as
potential instructions and label the remaining 34,822 sentences as
answers, requiring both segmentation and synthetic alignment.

\vspace{1mm}
\begin{table}[h]
\centering
\small
\setlength{\tabcolsep}{3pt}
\begin{tabular}{lcccc}
\toprule
\textbf{Dataset} & \textbf{Pairs Used} & \textbf{Unlab.\,Q} &
\textbf{Unlab.\,A}\\
\midrule
Alpaca–GPT4          & 20,000 & 10,000 & 10,000\\
Dolly-15k            & 15,000 & 7,500  & 7,500\\
Medical-Alpaca       & 20,000 & 10,000 & 10,000\\
\midrule
OASST-1 (logs)       & 40,000 & 15,126 & 24,874\\
WikiHow-4w (text)    & 40,000 & 5,178  & 34,822\\
\bottomrule
\end{tabular}
\caption{Statistics after subsampling.  “Unlab.” counts denote fragments
lacking the opposite side of the pair and hence requiring synthesis.}
\label{tab:data}
\end{table}

\paragraph{Baselines.}
We benchmark six settings—\textbf{Vanilla} (zero-shot), \textbf{All-SFT} (100 \% gold pairs), \textbf{80\%-SFT}, three seed-based back-translation variants(\textbf{Rand-$k$ \%}, which samples $k\!\in\!\{5,10,20\}$ \% of the gold pairs uniformly at random, and \textbf{Clust-$k$ \%}, which chooses the same proportion from K-means clusters), and our seed-free \textbf{Cycle-Inst/Filt}.  
Full implementation details are provided in Appendix~\ref{app:baselines}.

\subsection{Evaluation Protocol}
\label{ssec:eval}

\paragraph{Standard instruction metrics.}
For models trained on general instruction datasets, we follow \texttt{InstructEval} \citep{chia2023instructeval} and report accuracy (\%) on \textsc{MMLU} \citep{hendrycks2020measuring}, \textsc{BBH} \citep{suzgun2022challenging}, \textsc{CRASS} \citep{frohberg2021crass},
\textsc{DROP} \citep{dua2019drop}.
For the medical track we now follow the \emph{eight} specialised
sub-domains of the \textsc{MMLU} benchmark—%
CK (Clinical Knowledge), CB (College Biology),
CC (College Chemistry), CM (College Medicine),
HB (High-School Biology), HC (High-School Chemistry),
MG (Medical Genetics), and PM (Professional Medicine).

\paragraph{Open-ended quality.}
We further report \texttt{AlpacaEval} \citep{li2023alpacaeval} win-rate, the fraction of pairwise comparisons a system wins against the ALL-SFT baseline, providing a human-preference proxy for factual quality.

\subsection{Implementation Details}
\label{ssec:impl}

All experiments fine-tune the \textbf{Llama-3.1-8B-Base} \citep{grattafiori2024llama} checkpoint with
low-rank adaptation (LoRA) \citep{hu2022lora}. For LoRA, we set the rank to 8, the scaling factor $\alpha$ to 16, and apply a dropout rate of 0.05. The learning-rate schedule is cosine decay from an
initial value of $1\times10^{-4}$.  Training employs a micro-batch size
of $4$, an effective batch size of $32$, a sequence cutoff length of
$1024$ tokens, and runs for three epochs. All generation and evaluation are performed with vLLM \citep{kwon2023efficient}, using a maximum model length of 2048 tokens, top-k sampling with k = 10, a temperature of 0.2, and a generation limit of 500 tokens. Instruction-corpus experiments are trained on
\mbox{8 × NVIDIA RTX 3090} (24 GB each), whereas raw-document experiments
are trained on \mbox{8 × RTX 4090} (24 GB each).

\begin{table*}[t]
\centering
\small
\setlength{\tabcolsep}{4pt}
\begin{tabular}{l|c|ccccc|ccccc}
\toprule
 & & \multicolumn{5}{c|}{\textbf{OASST-1 (Dialogue Logs)}} & \multicolumn{5}{c}{\textbf{WikiHow-4w (Plain Text)}} \\
\cmidrule(r){3-7}\cmidrule(lr){8-12}
\textbf{Method} & \textbf{Annot.\,(\%)} & MMLU & BBH & CRASS & DROP & \textbf{Avg} & MMLU & BBH & CRASS & DROP & \textbf{Avg} \\
\midrule
Vanilla      & 0   & 55.85 & 37.16 & 59.48 & 36.01 & 47.13 & 55.85 & 37.16 & 59.48 & 36.01 & 47.13\\
\midrule
\multirow{3}{*}{Rand}
             & 5   & 56.99 & 37.48 & 65.69 & 36.99 & 49.29 & 57.64 & 36.59 & 68.61 & 34.72 & 49.39 \\
             & 10  & 57.10 & 37.85 & 63.87 & 38.14 & 49.24 & 57.90 & 37.25 & 64.96 & 35.34 & 48.86 \\
             & 20  & 57.38 & 36.32 & 70.44 & 38.87 & 50.75 & 57.86 & 37.29 & 68.25 & 35.13 & 49.63 \\
\midrule
\multirow{3}{*}{Clust}
             & 5   & 57.24 & 37.76 & 68.61 & 38.27 & 50.47 & 57.02 & 36.29 & 63.05 & 35.67 & 48.01 \\
             & 10  & 57.40 & 38.39 & 66.79 & 37.44 & 50.01 & 57.82 & 36.44 & 63.05 & 35.18 & 48.12 \\
             & 20  & 57.69 & 38.40 & 67.15 & 38.09 & 50.33 & 57.49 & 36.72 & 65.69 & 34.91 & 48.70 \\
\midrule
\textbf{Cycle-Inst (Ours)}   & 0   & 58.77 & 38.75 & \textbf{70.17} & 39.19 & 51.72 & 58.49 & 37.54 & 67.88 & 38.57 & 50.62 \\
\textbf{Cycle-Filt (Ours)}  & 0   & 59.07 & \textbf{38.98} & 70.07 & \textbf{39.52} & \textbf{51.98} & 58.70 & \textbf{38.50} & 67.52 & \textbf{39.54} & \textbf{51.07} \\
\midrule
\textcolor{DGray}{SFT-80}  & \textcolor{DGray}{80}  &
\textcolor{DGray}{58.81} & \textcolor{DGray}{38.26} & \textcolor{DGray}{68.61} & \textcolor{DGray}{37.80} & \textcolor{DGray}{50.87} &
\textcolor{DGray}{58.28} & \textcolor{DGray}{37.55} & \textcolor{DGray}{70.07} & \textcolor{DGray}{36.47} & \textcolor{DGray}{50.59} \\

\textcolor{DGray}{All-SFT} & \textcolor{DGray}{100} &
\textcolor{DGray}{\textbf{59.21}} & \textcolor{DGray}{38.56} & \textcolor{DGray}{70.07} & \textcolor{DGray}{38.92} & \textcolor{DGray}{51.69} &
\textcolor{DGray}{\textbf{58.83}} & \textcolor{DGray}{37.76} & \textcolor{DGray}{\textbf{71.90}} & \textcolor{DGray}{35.40} & \textcolor{DGray}{50.97} \\

\bottomrule
\end{tabular}
\caption{Results on OASST-1 and WikiHow for \textbf{Llama-3.1-8B}. Our methods beat all back-translation baselines and even achieve scores that exceed those of \textbf{All-SFT}.}
\label{tab:dialog-text-final}
\end{table*}

\section{Results}
\label{sec:results}

\subsection{Summary of Key Findings}
\label{ssec:summary}

\begin{figure}[ht]
  \centering
  \begin{subfigure}[t]{\linewidth}
    \centering
    \includegraphics[width=0.8\linewidth]{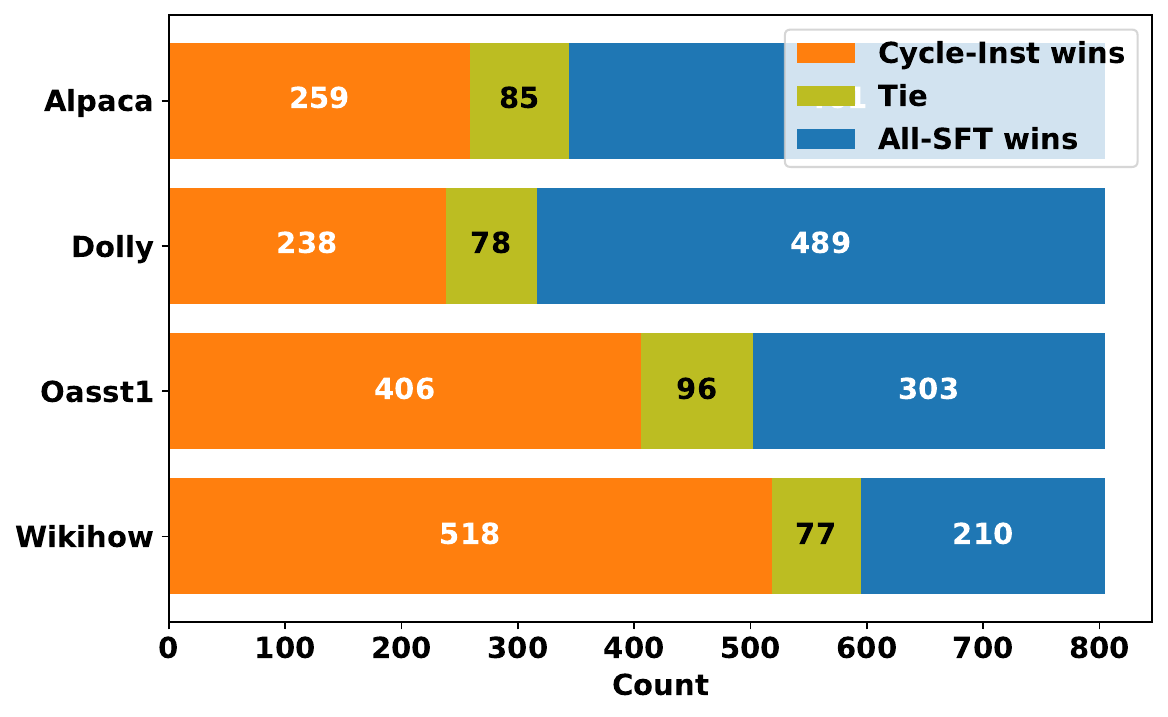} 
    \caption{Cycle-Inst vs. ALL-SFT.}
    \label{fig:recost_vs_full_alpaca}
  \end{subfigure}

  \vspace{1em} 

  \begin{subfigure}[t]{\linewidth}
    \centering
    \includegraphics[width=0.8\linewidth]{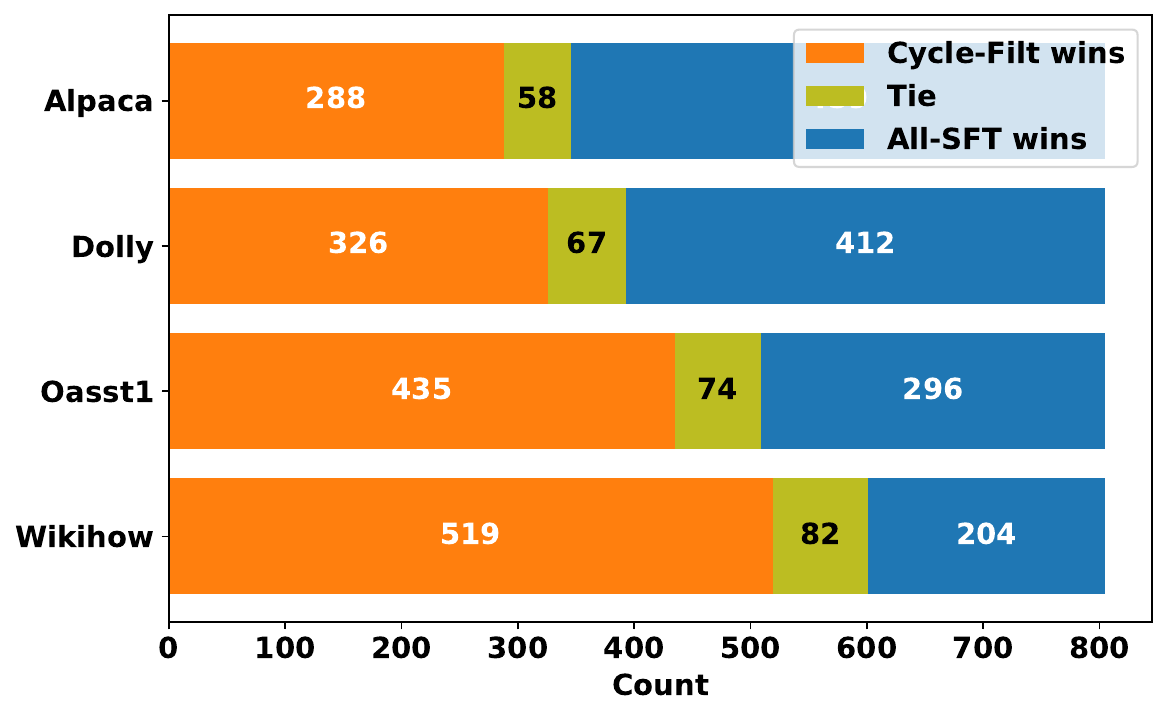} 
    \caption{Cycle-Filt vs. ALL-SFT.}
    \label{fig:recost_vs_full_alpaca_gpt4}
  \end{subfigure}

  \caption{Results on Alpaca Eval}
  \label{fig:recost_results}
\end{figure}

\textbf{1. Superior performance across all datasets.}  
Our primary focus is on \textbf{raw‐document‐to‐instruction‐tuning} data synthesis, where our method establishes a new \textbf{SOTA}.
On the nine medical \textsc{MMLU} sub-domains (Table~\ref{tab:medical-final}), \textbf{OASST-1} and \textbf{WikiHow-4w} (Table~\ref{tab:dialog-text-final}), \textsc{Cycle-Filt} achieves the highest scores among all methods.  
In each case it not only outperforms every seed-based back-translation baseline but also surpasses the fully supervised \textsc{All-SFT} model trained on 100\% of the data, demonstrating clear advantages when labels are scarce or absent.
On \textbf{Dolly-15k}, \textbf{Alpaca–GPT4} (Table~\ref{tab:general-final}), although our methods initially trail \textsc{All-SFT} (Table~\ref{tab:general-final}, Table~\ref{tab:dialog-text-final}), successive rounds of cycle-consistent pseudo-labeling allow \textsc{Cycle-Filt} to match—or in some metrics even marginally exceed—the performance of the model trained on 100\% of the data (for better models trained on \textbf{Dolly-15k} , see section~\ref{ssec:iterative-refinement}).
In every evaluation, our methods (\textsc{Cycle-Inst} and especially \textsc{Cycle-Filt}) outperform the supervised model trained on 80\% of the data. 
This holds true even when partial supervision is strong, underscoring the robustness of our synthetic-data approach.

\textbf{2. Stability and benefits of clustered seed selection.}  
Across seed budgets from 5 \% to 20 \%, \textsc{Clust-$k$} variants deliver stable, monotonic gains on both general and raw-text tasks, whereas \textsc{Rand-$k$} variants show erratic performance—occasionally plateauing or degrading. 
This emphasizes both the importance and challenges of a diversity-based seed data selection mechanism while also highlighting the \textbf{limitations} of back-translation approaches that generate data based on seed-data training.

\textbf{3. Effectiveness of cycle-consistency filtering.}  
Filtering via cycle consistency (\textsc{Cycle-Filt}) yields systematic improvements over the unfiltered \textsc{Cycle-Inst} across every benchmark, validating our hypothesis that noisy pseudo-pairs can be effectively removed through reconstruction verification.

\textbf{4. Addressing back-translation shortcomings on multi-task instruction augmentation.}  
All back-translation methods, initially underperform on the context-heavy Dolly-15k dataset (Table~\ref{tab:general-final}), trailing the fully supervised \textsc{All-SFT} model.
We believe this stems from Dolly’s multi-task instruction design: unlike purely declarative texts, its prompts specify explicit, diverse tasks that demand task-specific phrasing. 
Traditional back-translation focuses on matching answers to generated questions, yielding high QA alignment (see Table~\ref{tab:gpt4mini-eval}), but produces generic, one-size-fits-all instructions that lack the original task nuances (see Appendix~\ref{app:Dolly failure cases}).  
In contrast, our iterative pseudo-labeling framework gradually infuses task specificity: each cycle generates instructions increasingly tailored to the correct task category, mitigating the generic drift inherent to back-translation (see Appendix~\ref{app:Dolly_improvements_cases}). 
By the second iteration, our synthetic instructions recover the complexity of Dolly’s original prompts, significantly narrowing the performance gap and demonstrating robust adaptation even on complex, context-rich multi-task datasets (see Figure~\ref{fig:performance-iterations}).

\paragraph{Alpaca Evaluation.}
Figure~\ref{fig:recost_results} presents the Alpaca-based evaluation results on four datasets, comparing our two methods (\textsc{Cycle-Inst}, \textsc{Cycle-Filt}) against the fully supervised \textsc{All-SFT} baseline. 
In the open-ended evaluation, we see the same pattern as before: our methods outperform the fully supervised \textsc{All-SFT} when converting raw documents into instructions. 
However, when it comes to augmenting a general instruction-tuning pool, we fall slightly behind \textsc{All-SFT}. 
Moreover, \textsc{Cycle-Filt} consistently outperforms \textsc{Cycle-Inst} when pitted against \textsc{All-SFT}, further highlighting the superiority of our data-filtering strategy.

\begin{table}[t]
  \centering
  \small
  \setlength{\tabcolsep}{4pt}
  \scalebox{0.9}{%
    \begin{tabular}{lccccccc}
      \toprule
                     & \multicolumn{3}{c}{Random} & \multicolumn{3}{c}{Cluster} & Cycle-Inst \\
      \cmidrule(lr){2-4} \cmidrule(lr){5-7} \cmidrule(lr){8-8}
      Annot.\,(\%)   & 5     & 10    & 20    & 5      & 10      & 20      & 0           \\
      Alpaca         & 9.21  & 8.99  & 9.09  & 9.17   & 9.14    & 9.31    & \textbf{9.46} \\
      Dolly          & 9.54  & 9.48  & 9.15  & 9.19   & 9.50    & 9.27    & \textbf{9.90} \\
      MedAlpaca      & 9.80  & 9.88  & 9.89  & 9.88   & 9.89    & 9.80    & \textbf{9.96} \\
      OASST-1        & 8.45  & 8.39  & 8.53  & 8.64   & 8.39    & 8.54    & \textbf{8.75} \\
      WikiHow        & 9.14  & 9.04  & 9.17  & 8.91   & 9.15    & 9.09    & \textbf{9.43} \\
      \bottomrule
    \end{tabular}%
  }
  \caption{GPT-4o mini evaluation of synthetic QA pair alignment.}
  \label{tab:gpt4mini-eval}
\end{table}

\subsection{Synthetic Data Quality via \textsc{GPT-4o mini}}
\paragraph{GPT-4o Mini QA Pair Quality.}
To quantify how well each generated question matches its answer, we randomly sample 500 synthetic QA pairs per method and ask \textsc{GPT-4o mini} to assign a single relevance score on a 0–10 scale (See Appendix~\ref{PROMPT:GPT4OMINI_QAEvaluator}).
Table~\ref{tab:gpt4mini-eval} reports the average scores across four datasets. 
The results show that our methods yield substantially higher QA alignment than other back-translation baselines.

\paragraph{Correlation with Performance.}
Table~\ref{tab:pearson_results} shows the Pearson correlation coefficients between GPT-4o mini QA pair quality scores and downstream performance for each method and dataset. 
We observe consistently high correlation values across all methods and corpora, indicating a strong positive relationship between QA alignment quality and task performance. 
This result further validates our core hypothesis: by leveraging unlabeled text to learn a broader, more diverse data distribution, we can synthesize tightly paired instruction–text examples that substantially improve the overall quality of the generated training data.

\begin{table}[t]
  \centering
  \small
  \begin{tabular}{lcc}
    \toprule
    Dataset    & Pearson $r$ & $p$-value \\
    \midrule
    Alpaca  & 0.904       & 0.0052    \\
    Dolly  & 0.743       & 0.0559    \\
    MedAlpaca  & 0.646       & 0.117     \\
    OASST-1  & 0.872       & 0.0105    \\
    WikiHow  & 0.853       & 0.0146    \\
    \bottomrule
  \end{tabular}
  \caption{Pearson Correlation Coefficients and $p$-values between QA pair quality scores and downstream performance Across Datasets}
  \label{tab:pearson_results}
\end{table}

\subsection{Understanding of the Cycle-Instruct's Iterative Refinement}
\label{ssec:iterative-refinement} 

\begin{figure}[t]
  \centering
  \includegraphics[width=1.0\linewidth]{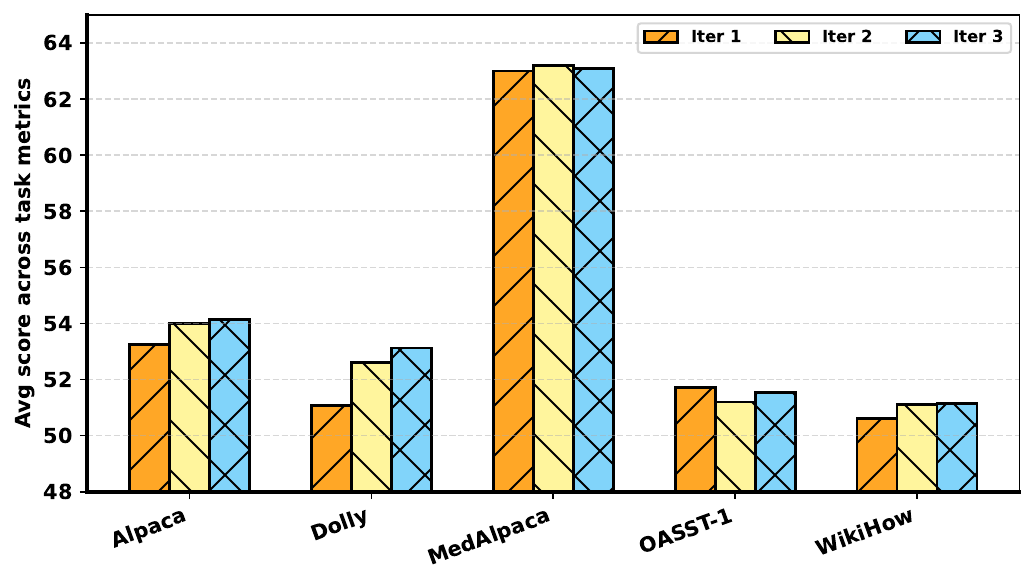}
  \caption{Cycle-Instruct's Performance over Iterations across different datasets.}
  \label{fig:performance-iterations}
\end{figure}

\paragraph{Performance over Iterations.}
Figure~\ref{fig:performance-iterations} illustrates the change in performance for our methods over three rounds of iteration on each dataset (see Appendix~\ref{app:iter_cyc} for more details). 
We observe that Dolly continues to improve steadily, Alpaca shows a modest uplift, whereas the other datasets plateau: MedAlpaca, OASST-1 and WikiHow remain essentially flat. 
As discussed in Section~\ref{ssec:summary}, this happens because on the simpler instruction datasets the model already generates sufficiently realistic pseudo-labels in the first training cycle, making further iterations unnecessary. 
Indeed, the strong performance gains seen after just one round in our main experiments further confirm that a single iteration is enough to capture the essential signal when task formats are straightforward.

%% file: latex/sections/appendix.tex
\newpage

\section{Related Work}
\label{sec:related-work}

\subsection{Supervised Instruction Tuning}

Supervised Instruction Tuning refers to fine-tuning large language models on explicit, human-written (or high-quality synthetic) instruction–response pairs that span a wide spectrum of tasks.  Initial research focused on established NLP benchmarks and demonstrated that models fine-tuned on these task collections could generalize zero-shot to novel text-based problems \citep{wei2021finetuned,sanh2021multitask,mishra2021cross}. With the advent of ChatGPT and other conversational systems, the community shifted toward general-purpose instruction corpora that cover reasoning, coding, multimodal description, and dialogue, laying the groundwork for assistant-style LLMs \citep{ouyang2022training,chiang2023vicuna,taori2023stanford}. In this supervised paradigm, the model first learns to map diverse prompts to high-quality answers, after which additional alignment steps such as RLHF or DPO can further refine helpfulness and safety \citep{ouyang2022training,rafailov2023direct}. Crowd-sourced efforts such as \textsc{OpenAssistant}~\citep{kopf2304openassistant} and \textsc{OpenHermes}~\citep{conover2023free} further democratise instruction data by collecting large human-annotated corpora. Despite strong performance, these approaches incur substantial annotation cost and remain constrained by the scope and biases of biases of available human prompts, motivating work on less supervised alternatives.


\subsection{Backtranslation-Based Instruction Tuning}
Instruction backtranslation generates synthetic instruction–response pairs for unlabeled text using a small seed set.  
Humpback~\citep{li2023self} repeatedly augments web documents with seed-conditioned prompts and self-curates the best candidates.  
\textsc{ReInstruct}~\citep{chen2024reinstruct} adds lightweight passage filtering and answer rewriting.  
\textsc{LongForm-C}~\citep{koksal2023longform} grounds answers in real documents before asking GPT-3 to author the corresponding instructions.  
\textsc{DoG-Instruct}~\citep{chen2023dog} wraps human-written documents into instruction form to curb hallucination while shrinking data size.

Although their pipelines and filtering strategies differ, all of these approaches ultimately derive data quality from the same backtranslation mechanism and thus inherit its common drawbacks: they still (i) depend on seed prompts or an external teacher model, (ii) suffer from data inefficiency because many synthetic pairs are low quality and must be filtered, and (iii) inherit distributional biases from the limited seed set, restricting instruction diversity. In contrast, \textsc{Cycle-Instruct} removes the seed bottleneck altogether: its dual self-training loop learns directly from raw corpora, needs no carefully selected seed data, and creates pseudo-pairs whose quality is enforced by cycle consistency rather than heavy post hoc filtering.  Our experiments therefore focus on comparing the \emph{generation paradigm} itself—seed-driven backtranslation versus fully seed-free cycle training.  The results show that \textsc{Cycle-Instruct} consistently delivers higher downstream accuracy and human preference, confirming the advantages of seed-free, cycle-consistent data synthesis over traditional backtranslation pipelines.

\section{Baseline Details}
\label{app:baselines}

\begin{itemize}[leftmargin=1.5em]
  \item \textbf{Vanilla.} Zero-shot performance of the base model (0 \% labels).

  \item \textbf{All-SFT.} Supervised fine-tuning on 100 \% gold instruction–response pairs.

  \item \textbf{80\%-SFT.} Supervised fine-tuning on 80 \% of the gold pairs.

  \item \textbf{Rand-$k$ \% / Clust-$k$ \%.}  
        To model the scarcity of seed examples, we sample only $k\!\in\!\{5,10,20\}$ \% of the gold pairs—uniformly at random (\textbf{Rand}) or via K-means clustering in embedding space (\textbf{Clust}).  
        The observed side of each seed pair is \emph{back-translated} to create its missing counterpart, and the model is fine-tuned on the resulting synthetic corpus.  
        Varying $k$ explores the trade-off between seed size and performance.

  \item \textbf{Cycle-Inst / Cycle-Filt.}  
        Our one-round, seed-free framework with 0 \% labels.  
        \textbf{Cycle-Inst} trains directly on all synthetic pairs produced by the dual self-training loop, whereas \textbf{Cycle-Filt} further prunes pairs whose reconstructions violate a 5 \% k-center-greedy cycle-consistency threshold (see Section~\ref{sec:cycle-consistency} for more details).
\end{itemize}

\section{Prompt Templates}
\label{app:prompts}
\input{latex/figure/full_prompts}

\section{Cases}
\subsection{Dolly Failure Cases}
\label{app:Dolly failure cases}

\begin{figure*}[t]
  \centering
  \includegraphics[width=\textwidth]{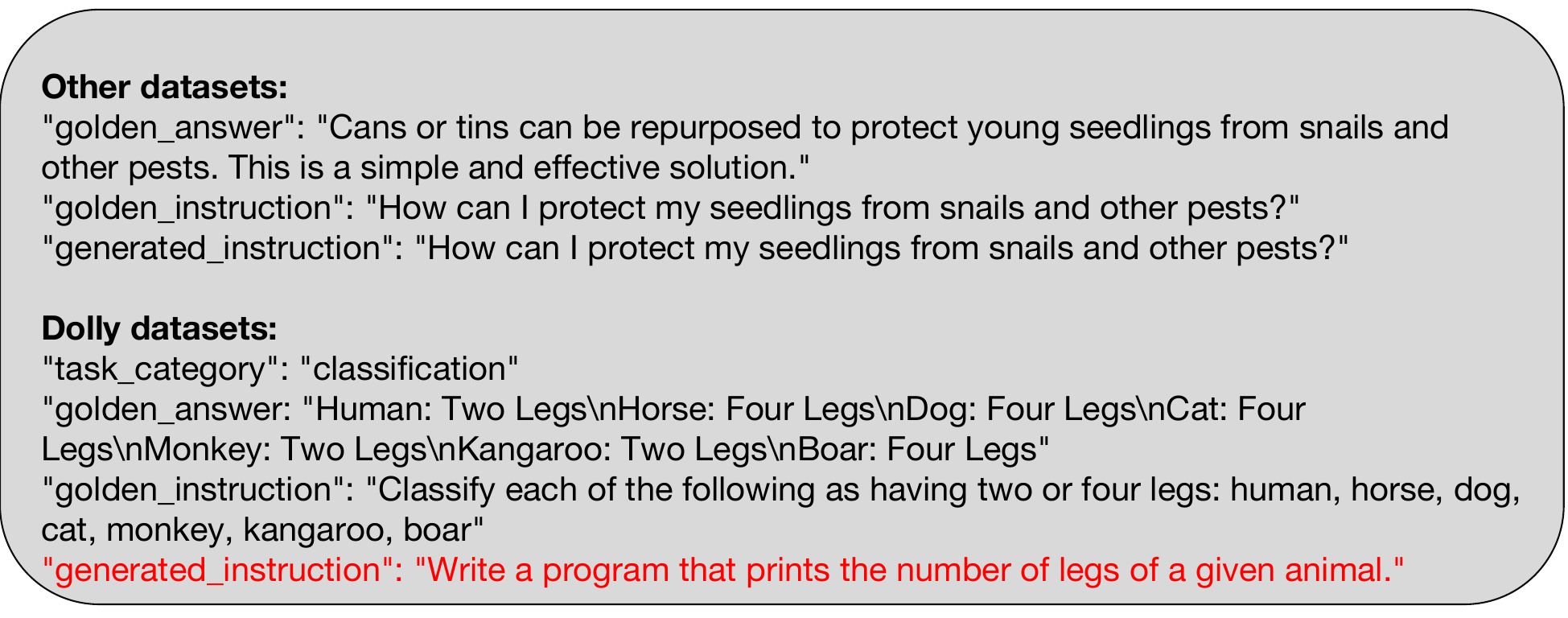}
  \caption{Unlike declarative texts in other datasets, Dolly’s multitask-based prompts embed explicit tasks, so the heightened variability and compatibility issues between prompts make accurate inversion particularly difficult and often lead to the generation of low-quality instructions.}
  \label{fig:Dolly failure cases}
\end{figure*}

Figure ~\ref{fig:Dolly failure cases} highlights how traditional back-translation succeeds on declarative web texts but fails on Dolly’s context‐rich, multi‐task prompts, producing generic instructions that lose task specificity.

\subsection{Iterative Instruction Refinement Cases for Dolly-15k}
\label{app:Dolly_improvements_cases}
\begin{figure*}[t]
  \centering
  \includegraphics[width=\textwidth]{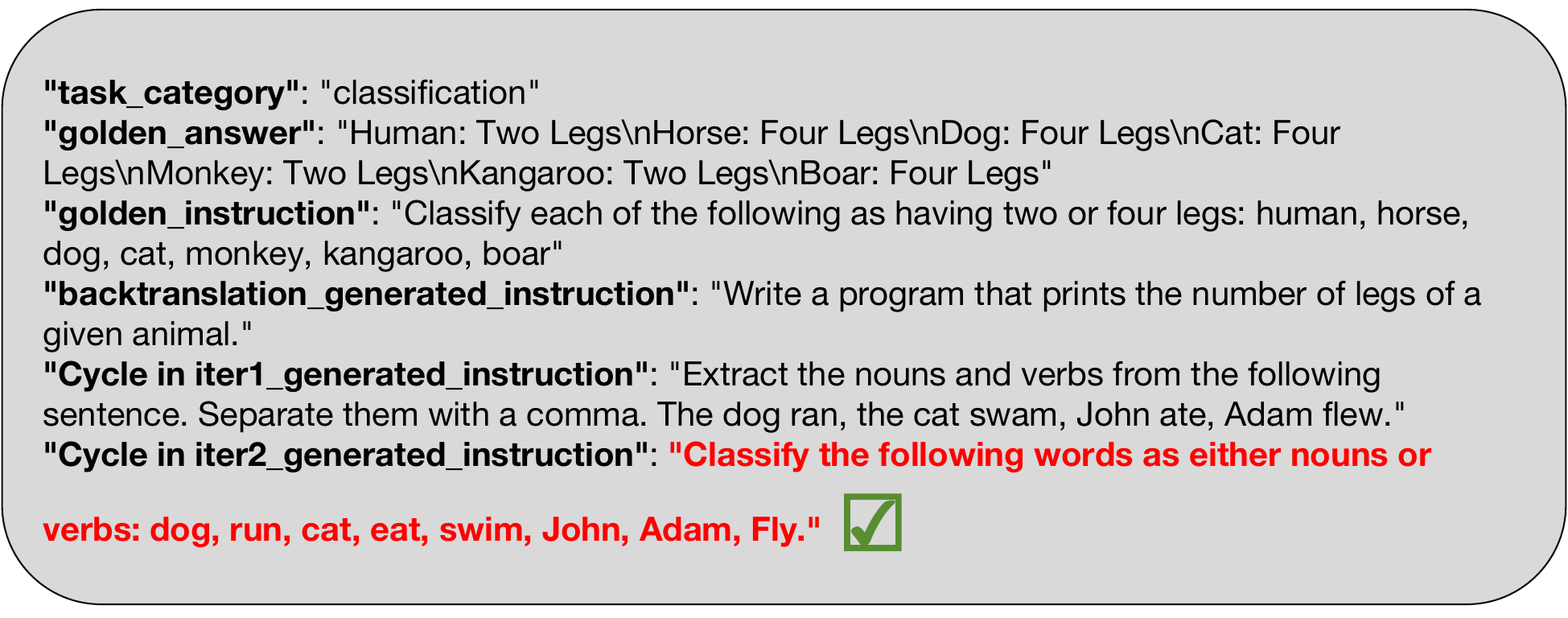}
  \caption{Iterative pseudo-instruction refinement for a classification example from Dolly-15k, showing the back-translation output, Cycle-Instruct iteration 1, and Cycle-Instruct iteration 2, progressively recovering the multi-instance classification task wording.}
  \label{fig:dolly_improvements_cases}
\end{figure*}

Figure~\ref{fig:dolly_improvements_cases} illustrates a classification example from the Dolly-15k dataset, contrasting the generic instruction produced by traditional back-translation with the progressively refined instructions generated by our Cycle-Filt framework over two iterations. This visualization demonstrates how iterative pseudo-labeling restores the original task‐specific phrasing.



\section{Iteration-wise Performance for Cycle-Inst}
\label{app:iter_cyc}

Below we report Table~\ref{tab:appendix_iter_general} and Table~\ref{tab:appendix_iter_medalpaca}, for each of five datasets, the Cycle-Inst method’s metrics over three self-training iterations. 

\begin{table*}[t]
  \centering\small
  \begin{tabular}{llccccc}
    \toprule
    \textbf{Dataset}     & \textbf{Iteration} & \textbf{MMLU} & \textbf{BBH} & \textbf{CRASS} & \textbf{DROP} & \textbf{Avg} \\
    \midrule
    Alpaca--GPT4         & Iter 1             & 59.01         & 39.28        & 74.82          & 39.86         & 53.24        \\
                         & Iter 2             & \textbf{59.39}         & 39.23        & \textbf{77.37}          & \textbf{40.05}         & 54.01        \\
                         & Iter 3             & 59.01         & \textbf{40.14}        & \textbf{77.37}          & 40.02         & \textbf{54.13}        \\
    \midrule
    Dolly-15k            & Iter 1             & 58.96         & 37.44        & 70.07          & 37.79         & 51.07        \\
                         & Iter 2             & 59.11         & 38.06        & 74.82          & \textbf{38.41}         & 52.60        \\
                         & Iter 3             & \textbf{59.13}         & \textbf{38.41}        & \textbf{76.64}          & 38.36         & \textbf{53.13}        \\
    \midrule
    OASST-1                & Iter 1             & 58.77         & \textbf{38.75}        & \textbf{70.17}          & 39.19         & \textbf{51.72}        \\
                         & Iter 2             & \textbf{59.15}         & 37.87        & 69.50          & 38.28         & 51.20        \\
                         & Iter 3             & 58.65         & 38.09        & \textbf{70.17}          & \textbf{39.28}         & 51.55        \\
    \midrule
    WikiHow               & Iter 1             & \textbf{58.49}         & 37.54        & 67.88          & \textbf{38.57}         & 50.62        \\
                    & Iter 2             & 58.29         & \textbf{38.37}        & 70.07          & 37.72         & 51.11        \\
                         & Iter 3             & 58.33         & 37.19        & \textbf{70.80}          & 38.27         & \textbf{51.15}        \\
    \bottomrule
  \end{tabular}
  \caption{Iteration-wise performance on four general datasets using Cycle-Inst.}
  \label{tab:appendix_iter_general}
\end{table*}

\begin{table*}[t]
  \centering\small
  \begin{tabular}{lccccccccc}
    \toprule
    \textbf{Iteration} & \textbf{CK} & \textbf{CB} & \textbf{CC} & \textbf{CM}
                      & \textbf{HB} & \textbf{HC} & \textbf{MG} & \textbf{PM}
                      & \textbf{Avg} \\
    \midrule
    Iter 1 & \textbf{0.668}       & 0.715       & \textbf{0.460}       & \textbf{0.578}
           & 0.742       & 0.473       & \textbf{0.760}       & 0.643
           & 0.630         \\
    Iter 2 & 0.657       & 0.722       & 0.440       & \textbf{0.578}
           & 0.755       & 0.478       & 0.750       & 0.673
           & \textbf{0.632}         \\
    Iter 3 & 0.638       & \textbf{0.729}       & 0.440       & 0.566
           & \textbf{0.758}       & \textbf{0.483}       & \textbf{0.760}       & \textbf{0.676}
           & 0.631         \\
    \bottomrule
  \end{tabular}
  \caption{Iteration-wise performance on Medical-Alpaca using Cycle-Inst.}
  \label{tab:appendix_iter_medalpaca}
\end{table*}

%% file: latex/figure/full_prompts.tex
\newtcolorbox{promptbox}[2][]{
  width=\textwidth,            
  boxrule=1.5pt,
  fontupper=\footnotesize,
  fonttitle=\bfseries\color{black},
  arc=5pt,
  rounded corners,
  colframe=black,
  colbacktitle=white!97!blue,
  colback=white!97!blue,
  title={#2},
  #1
}

\subsection{Prompt template for \texttt{REFORMAR\_PROMPTER}}
\label{PROMPTER}
\begin{figure*}[t]
  \centering
  \begin{promptbox}{REFORMAR\_PROMPTER}
Below is a block of Web text containing several paragraphs with question marks.  
Based on the content provided (or a portion of it), generate one plausible and clear question without summarizing the entire text. Maintain a natural, interrogative tone.

\textbf{Web text:}  
\{instruction\}

\textbf{Answer:}
  \end{promptbox}
  \caption{Prompt template for \texttt{REFORMAR\_PROMPTER}}
  \label{fig:prompt-REFORMAR_PROMPTER}
\end{figure*}

Figure~\ref{fig:prompt-REFORMAR_PROMPTER} presents the template used by the \texttt{REFORMAR\_PROMPTER}.  
Given a web passage that already contains question-like sentences, it prompts the model to craft one well-formed, natural question that could plausibly be asked about (part of) that passage, without collapsing the entire text into a summary. This question generation step seeds the subsequent answer-rewriting stage.

\subsection{Prompt template for \texttt{REFORMAR\_ASSISTANT}}
\label{ASSISTANT}
\begin{figure*}[t]
  \centering
  \begin{promptbox}{REFORMAR\_ASSISTANT}
A single-turn chat between a curious user and an artificial intelligence assistant.  
The assistant gives helpful answers to the user's questions.

\textbf{User:}  
Below is a block of Web text without any question marks.  
Please rewrite it into a fluent and coherent response that clearly conveys its intended meaning.  
The overall structure should remain similar, but ensure the language flows smoothly and the purpose is unmistakable.

\textbf{Web text:}  
\{output\}

\textbf{Assistant:}
  \end{promptbox}
  \caption{Prompt template for \texttt{REFORMAR\_ASSISTANT}.}
  \label{fig:prompt-REFORMAR_ASSISTANT}
\end{figure*}

Figure~\ref{fig:prompt-REFORMAR_ASSISTANT} presents the template used by the \texttt{REFORMAR\_ASSISTANT}. 
Here, the “assistant” is instructed to rewrite that passage into a direct answer that is fluent, coherent, and preserves the original structure—laying the groundwork for a clean \emph{(question, answer)} pair.

\subsection{Prompt template for \texttt{Pseudo-Answer Generation}}
\label{Pseudo-Answer Generation}
\begin{figure*}[t]
  \centering
  \begin{promptbox}{Pseudo-Answer Generation}
A single-turn chat between a curious user and an artificial intelligence assistant.  
The assistant gives helpful answers to the user's questions.

\textbf{User:}  
\{instruction\}

\textbf{Assistant:}
  \end{promptbox}
  \caption{Prompt template for \texttt{Pseudo-Answer Generation}.}
  \label{fig:prompt-Pseudo-Answer Generation}
\end{figure*}

Figure~\ref{fig:prompt-Pseudo-Answer Generation} depicts the template used when we already have an \emph{instruction} (or question) but need a synthetic answer.  
The model is cast as an assistant asked to provide a helpful, high-quality response, allowing us to bootstrap an answer in the absence of human annotations.

\subsection{Prompt template for \texttt{Pseudo-Instruction Generation}}
\label{Pseudo-Instruction Generation}
\begin{figure*}[t]
  \centering
  \begin{promptbox}{Pseudo-Instruction Generation}
Below is a reponse from an AI Assistant and its user instruction.  
The instruction is used as prompt for the response.

\textbf{Assistant:}  
\{output\}

\textbf{User:}
  \end{promptbox}
  \caption{Prompt template for \texttt{Pseudo-Instruction Generation}.}
  \label{fig:prompt-Pseudo-Instruction Generation}
\end{figure*}
Figure~\ref{fig:prompt-Pseudo-Instruction Generation} inverts the previous step:  
starting from a model-generated answer, it prompts the system to reconstruct a plausible \emph{instruction} that would elicit that answer.  
This back-translation closes the loop, enabling us to create balanced instruction–answer pairs even when only one side was originally available.

\subsection{Prompt template for \texttt{GPT4OMINI\_QA\_Evaluator}}
\label{PROMPT:GPT4OMINI_QAEvaluator}
\begin{figure*}[t]
  \centering
  \begin{promptbox}{GPT4OMINI\_QA\_Evaluator}
You are an AI evaluator. For a given Answer (A) and Generated Question (Q), score the relevance of the question to the answer on a scale from 0 to 10, where 0 means completely irrelevant and 10 means perfectly relevant. Respond ONLY with a single numeric score.

\textbf{A:} \{answer\}

\textbf{Q:} \{question\}

Relevance score (0–10):
  \end{promptbox}
  \caption{Prompt template for \texttt{GPT4OMINI\_QA\_Evaluator}.}
  \label{fig:prompt-GPT4OMINI_QAEvaluator}
\end{figure*}
Finally, Figure~\ref{fig:prompt-GPT4OMINI_QAEvaluator} illustrates the automated evaluation prompt.  
Given a synthetic \emph{(Q, A)} pair, GPT-4o-mini is asked to return a single relevance score from 0 – 10, enabling large-scale filtering of low-quality pairs without manual review.